# Saliency detection by aggregating complementary background template with optimization framework

*Chenxing Xia; Hanling Zhang; Xiuju Gao*

**Abstract:** This paper proposes an unsupervised bottom-up saliency detection approach by aggregating complementary background template with refinement. Feature vectors are extracted from each superpixel to cover regional color, contrast and texture information. By using these features, a coarse detection for salient region is realized based on background template achieved by different combinations of boundary regions instead of only treating four boundaries as background. Then, by ranking the relevance of the image nodes with foreground cues extracted from the former saliency map, we obtain an improved result. Finally, smoothing operation is utilized to refine the foreground-based saliency map to improve the contrast between salient and non-salient regions until a close to binary saliency map is reached. Experimental results show that the proposed algorithm generates more accurate saliency maps and performs favorably against the state-off-the-art saliency detection methods on four publicly available datasets.

## 1. Introduction

Saliency detection aimed at finding the most important part of an image [1, 2]. Numerous tasks in computer vision, such as image segmentation [3], object detection [4], object recognition [5], image retrieval [6], image and video compression [7], and content-aware image resizing [8] benefit from saliency detection as a pre-processing step to focus on the area of great importance.

Generally speaking, viewed from the information processing perspective, saliency algorithms can be categorized into either bottom-up (data and stimulus driven) or top-down (task and knowledge driven). Top-down algorithms [9-10] require supervised learning with manually labeled ground truth. To better distinguish saliency object from background, high-level information and supervised methods are incorporated to improve the accuracy of saliency map. Liu et al. [9] proposed a supervised approach to learn to detect a saliency region in an image. In [10], a dictionary learning is used to extract region features and Conditional Random Field (CRF) is employed to generate a saliency map. In contrast, bottom-up[11-14] saliency maps are formulated based on low-level cues such as edge, shape, and color. The appearance contrasts between objects and their surrounding regions, called contrast prior, is used in almost saliency algorithms. Besides contrast prior, more and more bottom-up methods prefer to construct the saliency map by choosing the image boundary as the background prior. This boundary prior is more general than previously used center prior, which assumes that the saliency object tend to appear near the image center. Admittedly, it is highly possible for the image border to be the background [15, 16]. However, if the object appears on the image boundary, the background prior will be imprecise and lead to the inaccurate results. To compensate, various modified methods have been proposed In [20], that the boundary which has the maximum Euclidean distance of any two of the four boundary-histogram in RGB three channels be removed and then the superpixels of the remaining three sides of the image be used as background queries. In [21], the image edge information was used to remove the foreground noise and obtain more stable and reliable background prior knowledge.

We propose a graph-based algorithm to improve the overall quality of the saliency maps in this paper. We first combine different border areas to form the background template. Specially, we use the nodes on each side of image as labelled background queries and compute the saliency of nodes based on their relevance to those queries. To solve the problem that some background regions are mistakenly highlighted as foreground, we product the foreground seeds by segmenting background-based saliency map(BBM) via adaptive threshold and compute a foreground-based saliency map(FBM). Finally, an iterative optimization framework is proposed to uniformly highlight the salient region and suppress the background region. The main steps of the proposed method are shown in Fig.1.

The contributions of our work include:

1) A self-adaptive weighted background template is proposed for saliency detection.

2) A new optimization framework is proposed for refining the saliency map that gives the test results when compared to other algorithms on three state-of-the-art public datasets.

The rest of this paper is organized as follows. In Section 2, we describe the details of graph construction. In Section 3 and Section 4, the estimation of background and foreground regions are described respectively. An iterative optimization framework is proposed to refine coarse saliency map in Section 5. Experimental results of the proposed algorithm and comparisons with several previous methods are presented in Section 6. Section 7 concludes this paper.

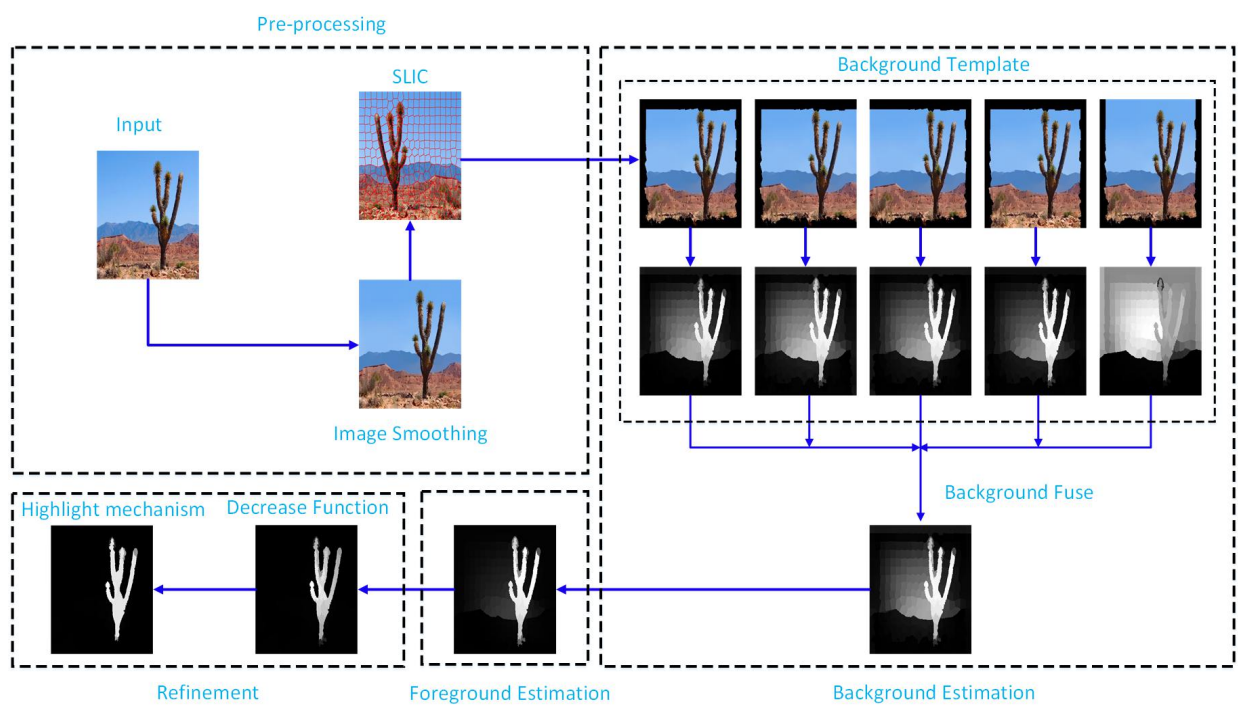

Fig.1 Main pipeline of the proposed saliency detection algorithm BTOF on an example image.

## 2. Related Work

Recently, numerous bottom-up saliency detection methods have been proposed, which prefer to generate the saliency map by utilizing the boundary information. Han et al.[11] proposed a novel saliency detection method using a new concept of optimal contrast by means of sparse coding principles. Afterwards, these hypotheses are compared using an entropy-based criterion and the optimal contrast is selected for building the saliency map. In [12], the contrast image boundary is used as a new regional feature vector to characterize the background. Yang et al. [13] rank similarity of image regions with foreground or background cues using a graph-based manifold ranking. The ranking is based on relevance of an element with respect to the given queries. A more robust boundary-based measure which takes the spatial layout of image patches into consideration is proposed in [14].

In addition, more graph-based approaches have gained great popularity due to the simplicity and efficiency of graph algorithms. Harel et al.[17] proposed the graph based visual saliency (GBVS) method, which multiple features are used to extract saliency information. Jiang et al.[12] introduced the discriminative regional feature integration (DRFI) method that integrates regional contrast, property and backgroundness descriptor together to formulate the master saliency map. Chang et al. [19] present a computational framework by constructing a graphical model to fuse saliency maps.

## 3. Graph Construction

(1) *Pre-Processing:* Xu et al. [22] make use of *L0* gradient minimization to sharpen major edges of images and eliminate a manageable degree of low-amplitude structures. To globally retain salient edges and suppress low-amplitude details, we employ the make method to smooth the original image. We also over-segment an input image into superpixels using the SLIC algorithm for computational efficiency [23]. The result is a set of compact superpixels that are homogenous in color and maintain image boundaries. N=200 superpixels is determined to the sufficient for detecting salient objects [13]. Define a graph G = (V, E) whose vertices V are superpixels and E are the connections of any two nodes in G quantified by a weight matrix $W = [w_{ij}]_{n \times n}$. The degree matrix $D = diag(d_{11},...,d_{nn})$ is then generated where $d_{ii} = \sum_j w_{ij}$.

(2) *Construct Edge Connections of Graph:* We subsequently add edges to the initial graph G to build a new graph model with the following rules: 1) Each node sharing edges is connected to each other. 2) Each node is connected to neighboring nodes neighboring it. 3) Any two nodes from the four boundaries of the graph are treated as connected.

(3) *Feature extracted:* This process is crucial to the estimation of the saliency map as the edge weights are calculated by comparing the feature descriptors of two nodes. A good feature vector should exhibit high contrast between foreground and background regions. In our work, we mainly adopt two kinds of features: color and texture. For color features, we consider mean color values in the CIELAB color space for each superpixel. For texture features, we utilize histogram of gradients (HOG). The HOG provides appearance features by using around the pixels' gradient information at fast speed implemented by Felzenszwalb et al. [29].

(4) *Construct Edge Weights of Graph:* The edge weights of graph encode the similarity between nodes. We define the distance between two superpixels as:

$$dist(c_i, c_j) = \alpha_1 \| v_i^{lab} - v_j^{lab} \| + \alpha_2 \chi^2 (h_i^{tex} - h_j^{tex}) \tag{1}$$

where $c = (v^{lab}, h^{tex})$ is the combined feature descriptor for superpixel i, $\alpha_1, \alpha_2$ are weighting parameters. Empirically, $\alpha_1$ and $\alpha_2$ in Eq. (1) are set to 0.6 and 0.4 respectively. $\chi^2(h_i - h_j) = \sum_{m=1}^{K} \frac{2(h_i(m) - h_j(m))^2}{h_i(m) + h_j(m)}$ is the chi-sequared distance between histogram $h_i$ and $h_j$ with K being the number of bins. With the constraints on edges, the weight between two nodes is defined by:

$$w_{ij} = \begin{cases} \exp(-\frac{dist(c_i, c_j)}{\sigma^2}) & c_i \text{ and } c_j \text{ are adjacent nodes} \\ 0 & others \end{cases} \quad (2)$$

where $c_i$ and $c_j$ denote the feature corresponding to two nodes i and j, and $\sigma$ is a constant that controls the strength of the weight, empirically $\sigma^2 = 0.1$. Then we get the affinity matrix W of the graph model G=(V, E).

## 4. Background Estimation (BBM)

As previously mentioned, it is possible that the foreground object be located at the image boundary. Using such a boundary as queries in the background will lead to wrong saliency estimation. We therefore propose a novel way to optimize the background prior before the background saliency estimation.

The set of boundary regions $B_i$ can be obtained by extracted from four boundaries of an image, which is defined as

$$B_i = [b_1^i, ..., b_k^i, ..., b_N^i] \quad (3)$$

where $B_i$ indicates the feature set of *i-th* boundary, $b_k^i$ denotes the feature of *k-th* superpixel in $B_i$, N is the total superpixel number and $i \in \{top, bottom, left, right\}$ indicates the four boundary location. We observe that the number of the foreground object near the image boundary is seldom more than two. Based on such discovery, we establishes the background template set $BT = \{BT_1, BT_2, BT_3, BT_4, BT_5\}$ by combining different boundary regions instead of only full-array to obtain more visual cues of the background model. A background template set based on the combination of boundary regions is shown in Table1. $B_i = 1$ and $B_i = 0$ indicate $B_i \in BT_k$ and $B_i \notin BT_k$ respectively.

Once the query boundaries are obtained, we can label the corresponding superpixels to be background. As a result, five saliency maps corresponding to $BT_1$ to $BT_5$ can be obtained. Taking $BT_1$ as an example, we use the nodes on four sides as the queries. More formally, we build a query vector $y = [y_1, y_2, ..., y_n]^T$, where $y_i = 1$ if node $\chi_i$ belongs to the queries, and $y_i = 0$ otherwise. Let $f$ be the ranking function assigning rank values $f = [f_1, ..., f_n]^T$ which could be obtained by solving the following minimization problem:

$$f^* = \arg\min_f \frac{1}{2} (\sum_{i,j=1}^n w_{ij} \| \frac{f_i}{\sqrt{d_i}} - \frac{f_j}{\sqrt{d_j}} \|^2 + \mu \sum_{i=1}^n \| f_i - y_i \|^2) \qquad (4)$$

where $\mu$ is a controlling parameter, which is set to 0.01. The optimized solution is given in [13,27,28] as:

$$f^* = (D - \alpha W)^{-1} y \qquad (5)$$

where $\alpha = 1/(1+\mu)$. Each element of the vector indicates the relevance of each node to the background queries, and its complement is the saliency measure. The vector is normalized to the range between 0 and 1, and the saliency map using $BT_1$ boundary prior, $S_{b1}$ can be written as:

$$S_{b1}(i) = \prod_{j \in BT_1}(1 - \overline{f_j^*}(i)) \qquad i = 1, 2, \ldots, N \qquad (6)$$

where $i$ indexes a superpixel node on graph, and $\overline{f_j^*}$ denotes the normalized vector. Similarly, we compute the other four maps $S_{b2}$, $S_{b3}$, $S_{b4}$ and $S_{b5}$ using the $BT_2$, $BT_3$, $BT_4$ and $BT_5$ as queries. The results are shown in Fig.2 (b-f)

Next, we treat the final saliency map as linear combination of the maps at individual compound mode, and obtain the weights in the linear combination by running a least-squares estimator over a validation dataset, indexed with Iv. Thus, our aggregated saliency map $S_{BBM}$ is formulated as follows,

$$S_{BBM} = \sum_{k=1}^{5} \lambda_k S_{bk} \qquad s.t. \{\lambda_k\}_{k=1}^{5} = \arg\min_{\lambda_1, \ldots, \lambda_5} \sum_{j \in I_v} \| S_{BBM}^j - \sum_k \lambda_k S_{bk}^j \| \qquad (7)$$

Table 1 A background template set based on the combination of boundary regions

| Template | $BT_1$ | $BT_2$ | $BT_3$ | $BT_4$ | $BT_5$ |
| --- | --- | --- | --- | --- | --- |
| $B_{top}$ | 1 | 1 | 1 | 1 | 0 |
| $B_{bottom}$ | 1 | 1 | 1 | 0 | 1 |
| $B_{letf}$ | 1 | 1 | 0 | 1 | 1 |
| $B_{right}$ | 1 | 0 | 1 | 1 | 1 |

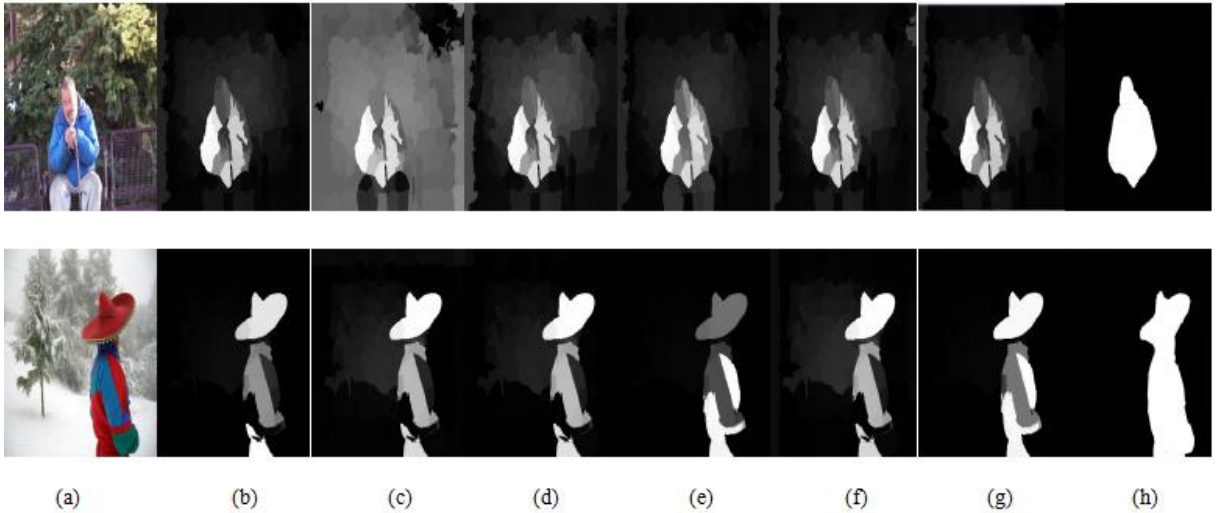

Fig.2 Comparison of saliency map using different combination of boundary nodes. (a) Original image. (b-f) Result of using $BT_1$, $BT_2$, $BT_3$, $BT_4$, $BT_5$ as boundary prior respectively. (g) Result of using our integration model. (f) Ground truth.

Fig.2 shows that the performance of our background model, Fig.2 (g) is better than others (Fig.2 (b-f)). We combine the advantages of the latter to form a new result, which suppresses background effectively. As shown the top row of Fig.2, while most salient region in an image can be highlighted from its surroundings by our background model, the background may not be adequately suppressed. To solve this problem, the saliency maps are further improved by the foreground cues as follows.

## 5. Foreground Estimation

Our background model can roughly estimate and separate the foreground object. However, it does not perform well when the background is suppressed deeply and the foreground is uniformly highlighted. Therefore, foreground-based saliency map is computed.

The BBM obtained in Section 3 is binary segmented using an adaptive threshold $T$, from which the nodes corresponding to salient region could be approximately separated. Threshold $T$ is computed as

$$T = a \cdot \min(S_{BBM}) + b \cdot \max(S_{BBM}) + c \cdot mean(S_{BBM}), \quad (8)$$

Where $a,b,c$ is regulatory factors. Fig.3 shows that when a=0.025, b=0.95, c=0.025, performance is stable. Using $T$, the foreground queries are obtained, Using defined indication vector $y$, the ranking function $\overset{++}{f}(i)$ can be calculated from Eq.5 as follow, which is treated as the FBM.

$$S_{FBM}(i) = \overset{++}{f}(i) \quad (9)$$

Where $i$ indexes a node of graph. The Fig.4 shows two examples where the backgrounds of the saliency maps are better suppressed, and the saliency region is almost uniformly highlighted.

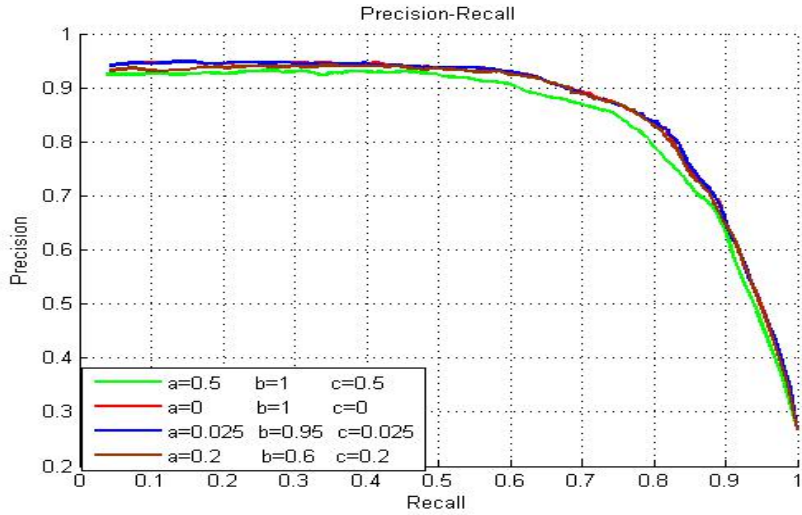

Fig.3 The parameter setting and performance of different threshold in the second stage on the SED1 dataset.

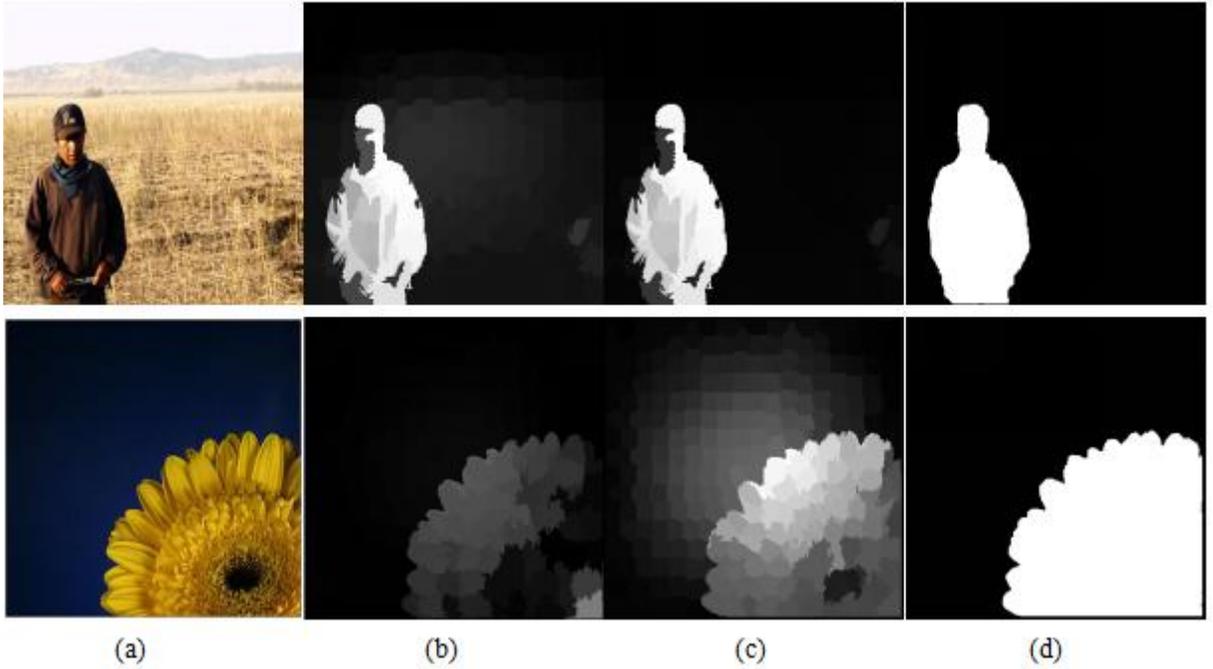

Fig.4 saliency map refined by the foreground cues. (a) Input image. (b) BMP. (c) FBM. (d) Ground truth.

## 6. Optimization Framework

While most saliency maps produced in Section 4 present good results, there are drawbacks in some saliency maps due to the complicated backgrounds of the original input images. As shown in the bottom row of Fig.4, backgrounds of the saliency maps are not suppressed adequately and the foreground does not obviously stand out from its surroundings. Therefore, we propose to improve the results via two-stage iterative optimization framework.

*Decrease function.* Salient objects can be of any size and be at any location, including image borders. Thus, for an image, location or size prior may do more harm than good. However, for the sets of images, the location prior can be beneficial ,because most of the current benchmarks are seriously biased, it helps to obtain better results on these datasets. Based on the $S_{FBM}$, we introduce a simple but effective superpixel-wise depression function to decrease the saliency of some regions in an image. A pixel-level saliency map is denoted by S(x, y) by assigning saliency value to each pixel of superpixel. To further suppress the background noises, we use a Gaussian smoothing kernel function, as follow

$$S(x,y) = S_{FBM}(x,y) * G(x,y) \tag{10}$$

$$G(x,y) = \exp(-(\frac{(x-x^c)^2}{2\sigma_x^2} + \frac{(y-y^c)^2}{2\sigma_y^2})) \tag{11}$$

where $x, y, x^c, y^c$ denote the coordinates of a pixel and the center coordinates of node i respectively. Empirically, we set $\sigma_x = 0.5W$ and $\sigma_y = 0.5H$, where W is the width and H is the height of an image. The region-level map is the average of pixel's values within a region.

$$S_{final1}(i) = \frac{1}{n_i} \sum_{(x,y) \in i} S(x,y) \tag{12}$$

Where $n_i$ indicates the number of pixels in node $i$. Fig.5 shows two examples where the background noise is suppressed by the decrease function.

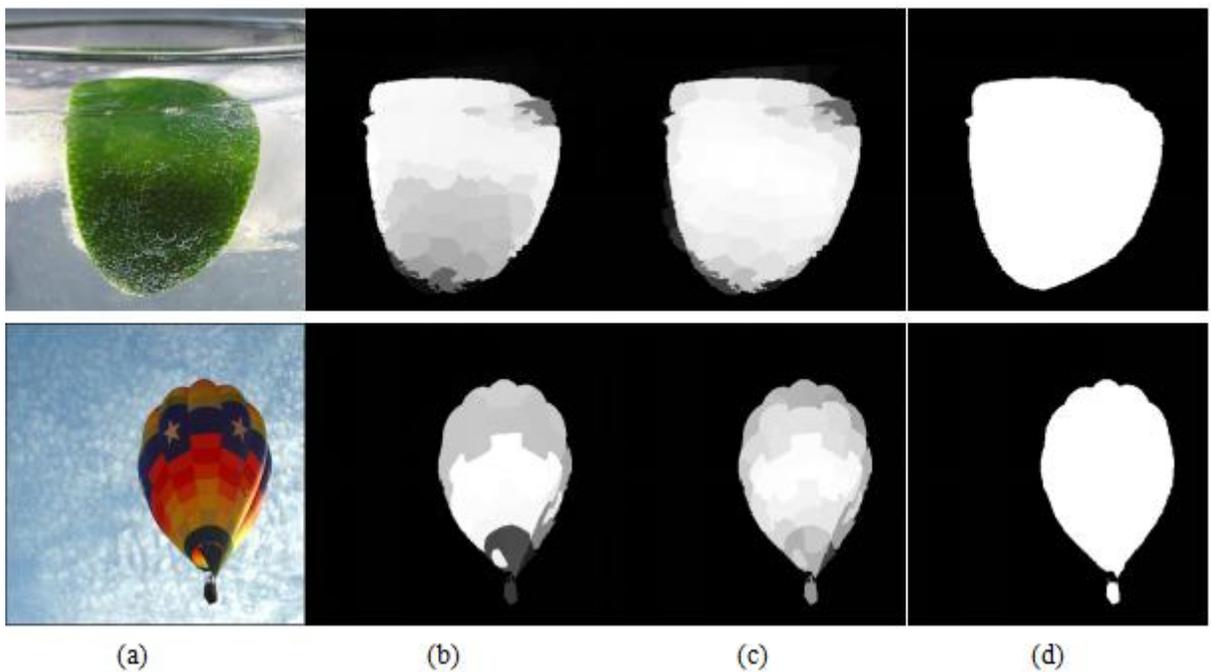

(a)　　　　　　(b)　　　　　　(c)　　　　　　(d)

Fig.5 Saliency map with diffusion mechanism. (a) Input image. (b) Saliency map without integration decrease function. (c) Saliency map with integration decrease function. (d) Ground truth.

***Highlight mechanism.*** It is known that the salient pixels are usually grouped together. In order to reduce background noisy and uniformly highlight the salient object, we exploit a highlight mechanism to refine the saliency map that the foreground of $S_{final1}$ are not effectively highlighted. At first, we group the nodes of the input image into *K* clusters by the K-means clustering algorithm. Suppose that there are *M* superpixels in cluster k (*k*=1,2,...,*K*). The modified saliency of node *i* is achieved by:

$$S_{fianl}(i) = \gamma_1 S_1(i) + \gamma_2 \frac{\sum_{j=1}^{M} weight(i,j) S_1(j)}{\sum_{j=1}^{M} weight(i,j)} \quad (13)$$

Where $weight(i,j) = \exp(-\frac{dist(i,j)}{\sigma^2})$ and $\gamma_1$ and $\gamma_2$ are weight parameters. The first term is the result of $S_{final1}$, while the second term is the weighted averaging saliency value of the other superpixels in the same cluster. The parameter $\gamma_1$ and $\gamma_2$ are empirically set to 0.5, and 8, which is sufficient to generate good saliency maps for most images. The examples in Fig. 6 show the effectiveness of the proposed method.

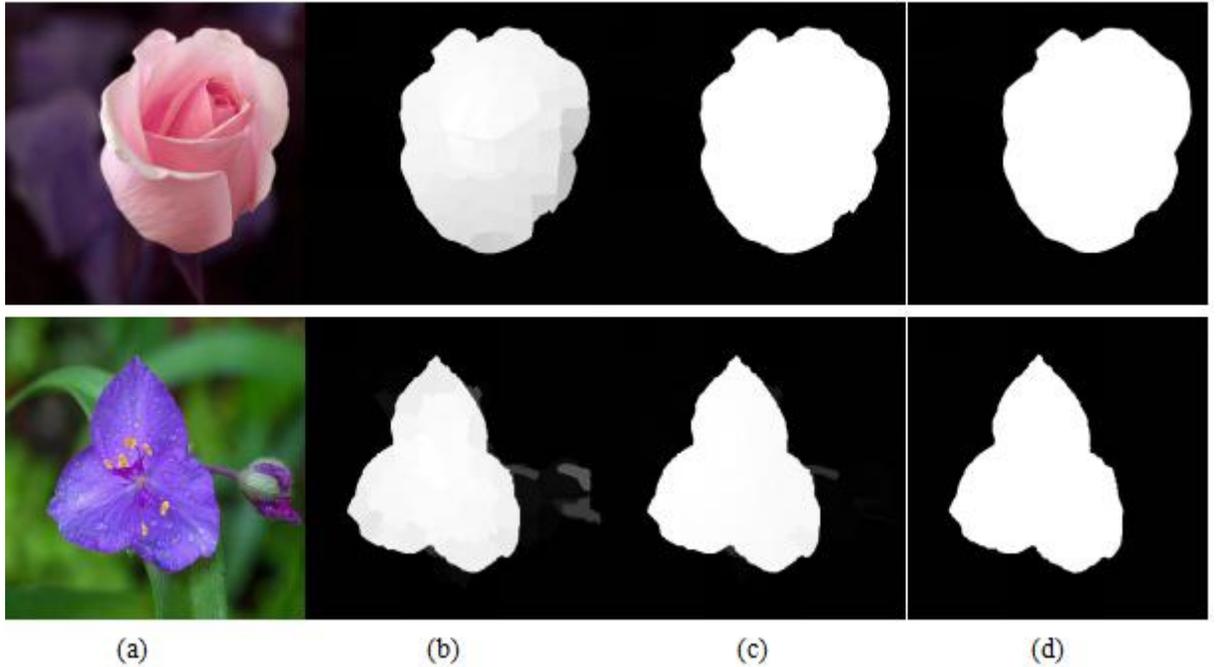

Fig.6 Saliency map with diffusion mechanism. (a) Input image. (b) diffused saliency map without integration highlight mechanism. (c) diffused saliency map with integration highlight mechanism. (d) Ground truth.

## 7. Experiment

To fully evaluate our algorithm, we conduct a series of experiments using four benchmark datasets involving various scenarios and include 12 recent solutions for comparison.

7.1 Experiment Setup

7.1.1 Datasets

We use four popular benchmark datasets to cover different scenarios. In particular, we use ASD[8] and DUT-OMRON [13] for images with a single salient object, iCoSeg [37] for cases with multiple salient object, and ECSSD [15] for images with complex scenes. The size and detailed characteristic of these benchmark datasets are presented in Tab. 1.

| Name | Size | Characteristics |
|------|------|-----------------|
| ASD[8] | 1000(imgs) | Single object, collected from MSRA[36], simple background, high contrast |
| DUT-OMRON [13] | 5168 | Single object, relatively complex background, more challenging |
| iCoSeg[37] | 643 | Multiple objects, various number of objects with different sizes |
| ECSSD[15] | 1000 | Structurally complex natural images, various object categories |

7.1.2 Salient object detection algorithms

The proposed salient object detection algorithm is compared with 12 state-of-the-art solutions as follows: GS[30], HS[31], LPS[32], MB[34], MC[33], MR[13], SF[26], PCA[35], BL[24], BSCA[25], RR[38], ST[39].

7.1.3 Evaluation metrics

For comprehensive evaluation, we use seven metrics including the precision-recall (PR) curve, the F-measure, the receiver operating characteristic (ROC) curve, area under the ROC curve (AUC), mean absolute error (MAE), overlapping ratio (OR) and the weighted F-measure (WF) score. Precision is defined as the percentage of salient pixels correctly assigned, while recall is the ratio of correctly detected salient pixels to all true salient pixels. F-measure is a weighted harmonic mean of precision (P) and recall (R): $F_\beta = \frac{(1+\beta^2) \cdot P \cdot R}{\beta^2 \cdot P + R}$, where we see $\beta^2 = 0.3$ to emphasize the precision[26]. We compute the precision, recall and F-measure with an adaptive threshold proposed in [1], which is defined as twice the mean saliency of the image. The ROC curve is generated from true positive rates and false positive rates which are obtained when we calculate the PR curve.

Although commonly used, the above metrics ignore the effects of correct assignment of non-salient pixels and the importance of complete detection. We therefore introduce the MAE and OR metrics to address these issues. Given a continuous saliency map S and the binary ground truth G, MAE is defined as the mean absolute difference between S and G : $MAE = mean(|S - G|)$ [26]. OR is defined as the overlapping ratio between the segmented object mask S' and ground truth G: $OR = |S' \cap G| / |S \cap G|$, where S' is obtained by binarizing S using an adaptive threshold, i.e., twice the mean values of S as in [18]. Finally, we adopt the recently proposed weighted F-measure (WF) metric [40], which is a weighted version of the traditional F-measure. It amends the interpolation, dependency and equal importance flaws of currently-used measures.

7.2 Comparison with State-of-the-Arts

The proposed BTOF algorithm is evaluated on the four benchmark datasets and compared with 12 recently proposed algorithms. The results are summarized in Tab. 2 and Fig. 7. Besides, Fig. 8 shows some qualitative comparisons. The results show that, in most cases, BTOF ranks first or second on the four benchmark datasets across different criteria. It is worth noting that, although ST [39] is the best performing method, it is a supervised one requiring a large amount of training. In contrast, our method is an unsupervised one, which skips the training process and therefore enjoys more flexibility.

TABLE 2  Results on four datasets in terms of WF, AUC, OR and MAE.

| | Metric | Our | MB | MR | PCA | MC | BSCA | LPS | GS | SF | HS | RR | BL | ST |
|---|---|---|---|---|---|---|---|---|---|---|---|---|---|---|
| (A) ASD | WF ↑ | 0.7598 | 0.71 | 0.0751 | 0.5087 | 0.6733 | 0.7029 | 0.7549 | 0.6494 | 0.5429 | 0.6491 | 0.7506 | 0.5937 | 0.7166 |
| | OR ↑ | 0.814 | 0.715 | 0.8101 | 0.6692 | 0.8115 | 0.8022 | 0.7924 | 0.7531 | 0.6601 | 0.7751 | 0.8131 | 0.7983 | 0.8273 |
| | AUC ↑ | 0.8773 | 0.8554 | 0.8632 | 0.8758 | 0.8758 | 0.8744 | 0.8541 | 0.8768 | 0.7968 | 0.871 | 0.8662 | 0.8837 | 0.8776 |
| | MAE ↓ | 0.0715 | 0.0908 | 0.0751 | 0.1561 | 0.0933 | 0.0858 | 0.0716 | 0.1073 | 0.1293 | 0.1109 | 0.0738 | 0.1291 | 0.0869 |
| | Metric | Our | MB | MR | PCA | MC | BSCA | LPS | GS | SF | HS | RR | BL | ST |
| (B) DUT | WF ↑ | 0.4058 | 0.4193 | 0.3787 | 0.2866 | 0.34672 | 0.3704 | 0.3609 | 0.3631 | 0.2709 | 0.3496 | 0.3839 | 0.3168 | 0.3866 |
| | OR ↑ | 0.4169 | 0.3924 | 0.4194 | 0.3413 | 0.4253 | 0.4094 | 0.3819 | 0.3718 | 0.3019 | 0.2274 | 0.4225 | 0.3967 | 0.4373 |
| | AUC ↑ | 0.8214 | 0.793 | 0.7813 | 0.8269 | 0.8203 | 0.8082 | 0.7671 | 0.7981 | 0.6678 | 0.8011 | 0.7818 | 0.8155 | 0.814 |
| | MAE ↓ | 0.1632 | 0.1566 | 0.1875 | 0.2065 | 0.1826 | 0.1907 | 0.187 | 0.1732 | 0.1468 | 0.2274 | 0.1845 | 0.2401 | 0.1815 |
| | Metric | Our | MB | MR | PCA | MC | BSCA | LPS | GS | SF | HS | RR | BL | ST |
| (C) iCoSeg | WF ↑ | 0.5625 | 0.5434 | 0.5452 | 0.4067 | 0.4606 | 0.5268 | 0.5325 | 0.5186 | 0.3959 | 0.5356 | 0.5538 | 0.4691 | 0.5456 |
| | OR ↑ | 0.5917 | 0.5085 | 0.5504 | 0.4272 | 0.5432 | 0.5518 | 0.5376 | 0.5196 | 0.465 | 0.5375 | 0.5772 | 0.5481 | 0.5597 |
| | AUC ↑ | 0.8046 | 0.7803 | 0.8062 | 0.7983 | 0.8071 | 0.8187 | 0.7996 | 0.8195 | 0.7284 | 0.8124 | 0.7971 | 0.8409 | 0.8313 |
| | MAE ↓ | 0.1536 | 0.163 | 0.1591 | 0.2007 | 0.1787 | 0.1601 | 0.1628 | 0.1672 | 0.1841 | 0.1757 | 0.1619 | 0.1831 | 0.1561 |
| | Metric | Our | MB | MR | PCA | MC | BSCA | LPS | GS | SF | HS | RR | BL | ST |
| (D) ECSSD | WF ↑ | 0.5295 | 0.5288 | 0.4931 | 0.3642 | 0.4545 | 0.513 | 0.4736 | 0.45 | 0.2588 | 0.4544 | 0.5003 | 0.4588 | 0.5119 |
| | OR ↑ | 0.5391 | 0.4966 | 0.5205 | 0.3951 | 0.5309 | 0.5491 | 0.4978 | 0.4606 | 0.3086 | 0.458 | 0.5314 | 0.5319 | 0.5495 |
| | AUC ↑ | 0.8043 | 0.7833 | 0.7899 | 0.7914 | 0.8161 | 0.8152 | 0.7825 | 0.7889 | 0.6244 | 0.8006 | 0.7959 | 0.8173 | 0.8111 |
| | MAE ↓ | 0.1719 | 0.1741 | 0.1892 | 0.2469 | 0.2023 | 0.1824 | 0.1959 | 0.2058 | 0.2187 | 0.2275 | 0.1837 | 0.2169 | 0.1903 |

7.2.1 Results on single-object images

The test on images with a single object is conducted on the ASD[8] and DUT-OMRON [13] datasets. The PR and F-measure are shown in the first two rows of Fig. 7, and the WF, OR, AUC and MAE scores in Tab. 2(A and B).

On ASD (Tab. 2(A)), BTOF achieves the best performance in terms of WF and MAE, and the second best in terms of OR and AUC. In the F-measure (the first row of Fig. 7), BTOF is the best two among those competitive methods. In the PR curve, BTOF is superior, as it achieves relatively good results over a large range.

On DUT-OMRON (Tab. 2(B)), all the methods perform worse than on ASD due to the large diversity and complexity of DUT-OMRON. BTOF performs the second or third best in terms of WF, AUC and MAE, with a very minor margin to the best results. The best MAE and OR scores are achieved by ST and SF. In the PR curves , the precision of BTOF is less impressive at high recall rates, but it is competitive a large range.

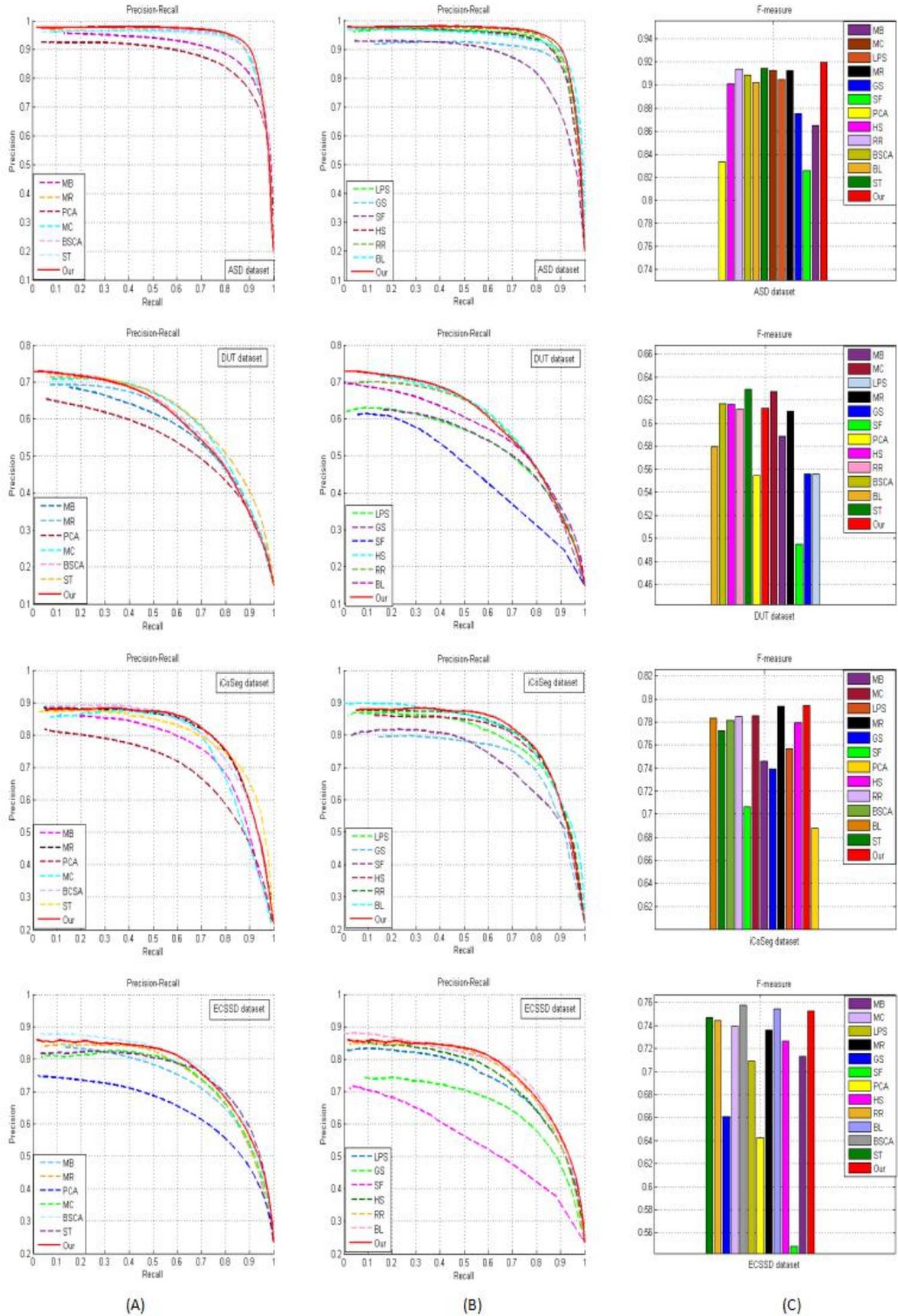

Fig.7.Quantitative comparison on four datasets in terms of PR and F-measure curves

7.2.2 Results on multiple-object images

Experiments on images with multiple salient objects are conducted on iCoSeg [37]. The PR and F-measure curves are shown in Fig. 7(the third row), and the WF, OR, AUC and MAE scores in Tab. 2(C ).

On iCoSeg (Tab. 2(C)), BTOF achieves the best performance in terms of WF, OR and MAE. The best AUC score are achieved by BL. Fig. 7(the third row), shows that the PR and F-measure curves of BTOF are superior or comparable to other methods.

7.2.3 Results on complex scene images

Our last comparison with the competing methods is conducted on ECSSD [15], which is known to involve complex scenes. As reported in Tab. 2(D), BTOF obtains the best performance in terms of WF and MAE, the third best in OR. According to Fig. 7(the last row), the PR curve of BTOF is the second best among those methods, while the area under the F-measure curve is the best. These results validate BTOF 'strong potential in handling images with complex scenes.

7.2.4 Visual comparison

Fig. 8 shows some visual comparisons of the best methods in the experiments. For single-object images, BTOF accurately extracts the entire salient object with few scattered patches, and assigns nearly uniform saliency values to all patches within the salient objects. For images with multiple objects, some methods (e.g., MC, MR and HS) miss detecting parts of the objects, while some (e.g., BSCA, GS and SF) incorrectly include background regions into detection results. By contrast,

BTOF pops out all the salient objects successfully. For the images with complex scenes, most methods fail to identify the salient objects, while BTOF locates them with decent accuracy. These results illustrate the robustness of the BTOF algorithm, and confirm the effectiveness of the proposed method.

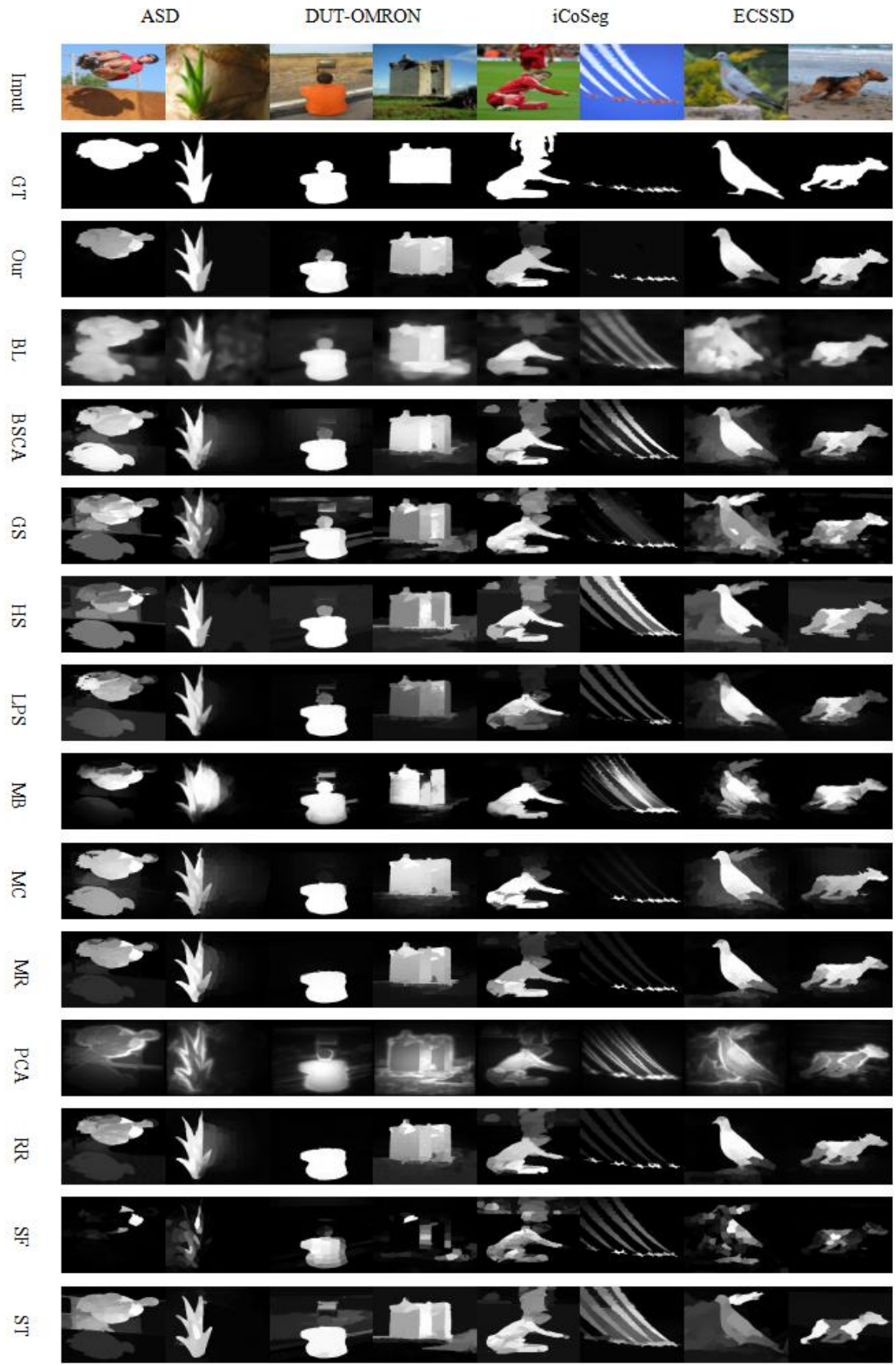

Fig.8 Visual comparisons of saliency maps of the best methods.

7.3 Analysis of components in the proposed model

To further understand the effects of the components in the proposed algorithm. Every stage in the proposed algorithm contributes to the final saliency map. In order to evaluate the property of our designing, we take a simple comparison among three stage through the precision-recall curve. Fig.9 shows the result of each step on ASD dataset, where the red line represents the first stage result, the black line denotes the second stage result and the green line refers to the three stage result. Based on the observations above, both the erroneous corner removal and the refinement have contributions to the overall performance: every stage operation goes foreword one by one.

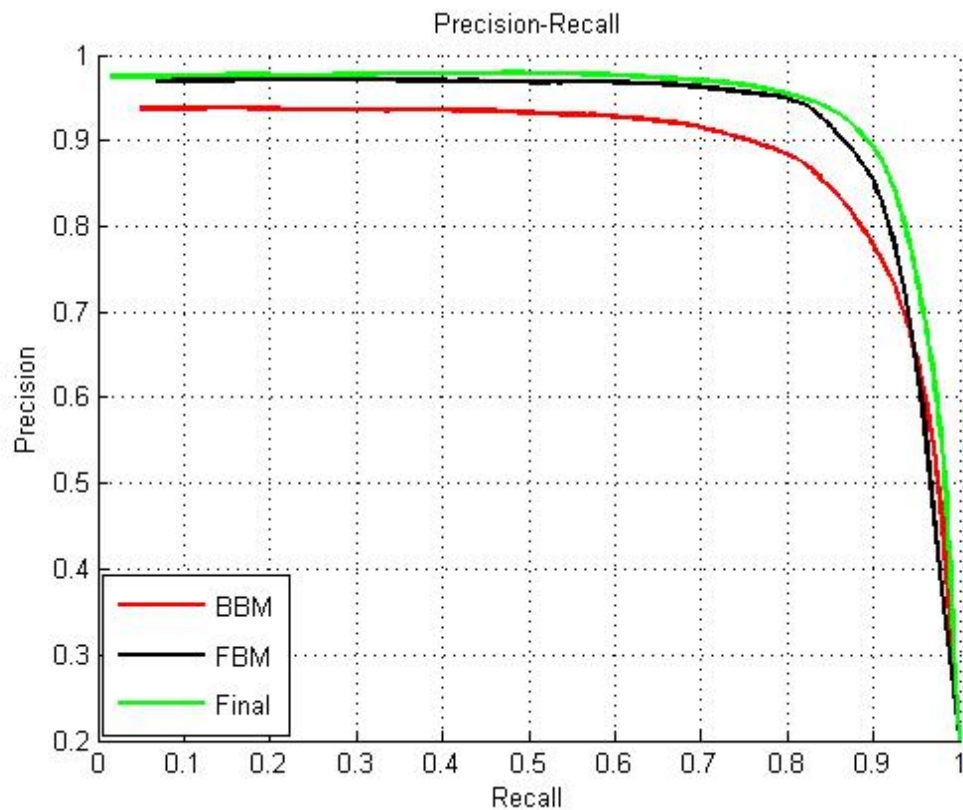

Fig.9 shows performance of each component in the proposed method on the ASD dataset.

## 7.4 Limitation and Analysis

Our model performs favorably against existing algorithms with higher precision and recall. However, as the BBM based on boundary template which may be insufficient in some scenarios and the FBM based on the former which may be unsafe if the former information is not accurate, hence, the proposed method does not work well if the BBM is not well. Fig.10 lists failure cases of our saliency map models. At the same time, we believe that investigating more sophisticated feature representations for our algorithm would be greatly beneficial. It would also be interesting to exploit top-down and category-independent semantic information to enhance the current results. There are, however, topics for further investigated.

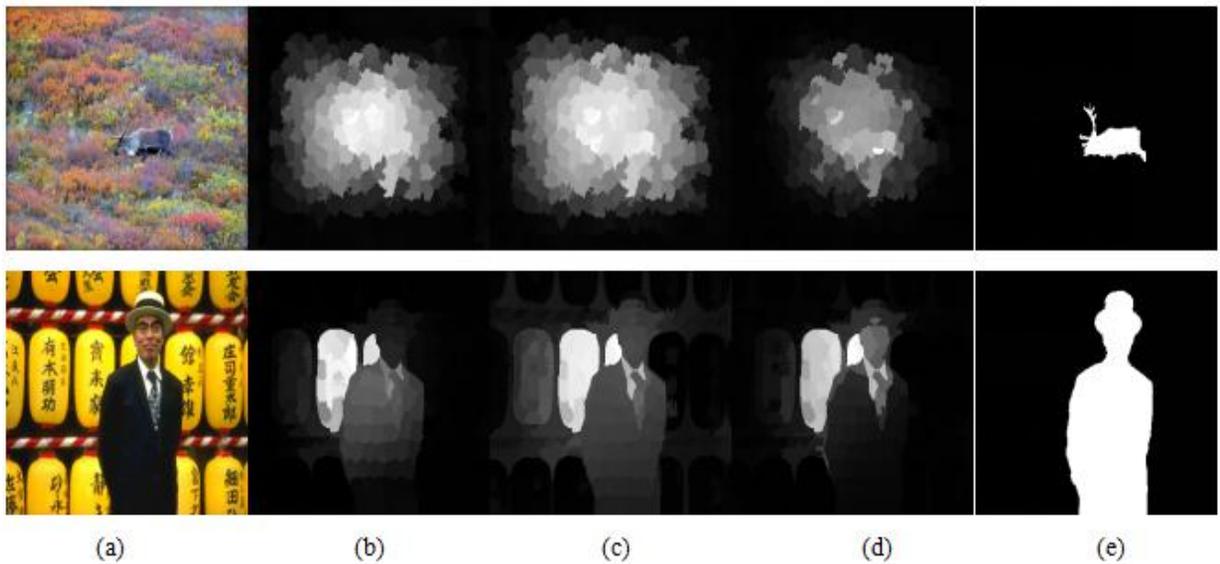

Fig.10 limitation. (a) Input. (b) Saliency map generated by background-based. (c) Saliency map generated by foreground-based. (d) Our final saliency map. (e) Ground truth.

# 8. Conclusion

We present a novel unsupervised saliency estimation method based on a novel background priors and smoothness. The robustness of background priors makes it especially useful for high accuracy background detection. The proposed feature distance metrics effectively and efficiently combines color and texture cues to represent the intrinsic manifold structure. Finally, a series of refinement techniques are applied to region-based saliency detection, which prove that the proposed approach can effectively improve the results and achieve the start-of-the-art performance.

**Acknowledgment**

This work was supported by the National Natural Science Foundation of China (Grant No.61672222), the Key Science and Technology Planning Project of Hunan province, China(Grant No.2014GK2007), the Priority Academic Program Development of Jiangsu Higer Education Institutions；Jiangsu Collaborative Innovation Center on Atmospheric Environment and Equipment Technology. We also like to acknowledge helpful comments from Prof. Tak-Shing Yum.